\documentclass{styles/svproc}
\usepackage{url}

\usepackage[utf8]{inputenc}
\usepackage[table,xcdraw]{xcolor}
\usepackage{soul}
\usepackage{filecontents,lipsum}
\usepackage{enumitem}
\usepackage{subcaption}
\usepackage{amsmath}
\usepackage{amssymb}
\usepackage{comment}
\usepackage{units}
\usepackage{xspace}

\usepackage{graphicx}

\newcommand{\IG}{\ensuremath{\text{IG}}\xspace}
\begin{document}
\mainmatter              
\title{Monitoring and mapping of crop fields with UAV swarms based on information gain}
\titlerunning{Information-gain based mapping with UAV swarms}  
%
\author{Carlos Carbone\inst{1} \and Dario Albani\inst{1,2,7} \and Federico Magistri\inst{3} \and Dimitri Ognibene\inst{5,6} \and Cyrill Stachniss\inst{3} \and Gert Kootstra\inst{4} \and 
Daniele Nardi\inst{1} \and Vito Trianni \inst{2}}
\authorrunning{Carlos Carbone et al.} 
%
%
\institute{%
DIAG, Sapienza University of Rome, Italy\\
\email{carbone@uniroma1.it} 
\and
ISTC, National Research Council, Rome, Italy\\
\and
University of Bonn, Germany
\and
Wagening University and Research, The Netherlands
\and
University of Milano-Bicocca, Milan, Italy
\and
University of Essex, Colchester, UK
\and
Technology Innovation Institute, Abu Dhabi, UAE
}

\maketitle              

\begin{abstract} 
Monitoring crop fields to map features like weeds can be efficiently performed with unmanned aerial vehicles (UAVs) that can cover large areas in a short time due to their privileged perspective and motion speed. However, the need for high-resolution images for precise classification of features (e.g., detecting even the smallest weeds in the field) contrasts with the limited payload and flight time of current UAVs. Thus, it requires several flights to cover a large field uniformly. However, the assumption that the whole field must be observed with the same precision is unnecessary when features are heterogeneously distributed, like weeds appearing in patches over the field. In this case, an adaptive approach that focuses only on relevant areas can perform better, especially when multiple UAVs are employed simultaneously. Leveraging on a swarm-robotics approach, we propose a monitoring and mapping strategy that adaptively chooses the target areas based on the expected information gain, which measures the potential for uncertainty reduction due to further observations. The proposed strategy scales well with group size and leads to smaller mapping errors than optimal pre-planned monitoring approaches.

\end{abstract}
\section{Introduction}

Precision farming requires high-quality data from the field in order to support operational and strategic decisions~\cite{King:2017bf}. In this respect, unmanned aerial vehicles~(UAVs) present a very flexible tool for remote sensing, as they can be deployed on-demand and can quickly monitor large areas~\cite{Zhang:2012kqa}. However, the platform limitations in terms of payload and flight time are often constraining, limiting the resolution of the acquired images and requiring multiple flights to cover extensive fields. 
At the same time, the heterogeneity of agricultural fields often requires high-resolution data only in certain areas where relevant features are present. In contrast, less interesting areas could be monitored on a coarser resolution. This is often the case, for instance, for weed-management practices, in which the density of weeds is not uniform across the field, as weeds appear in patches. To support efficient management of the field, a greater effort should be dedicated to those areas where relevant features are actually present. In this way, the limited energy budget of each UAV is properly allocated, avoiding dissipating it by flying over uninteresting areas. Given that the feature distribution over the field is a priori unknown, the monitoring and mapping strategies should be adaptive, responding to the observed field features in order to determine the best flight pattern, therefore ruling out predefined mission plans that do not take into account the actual distribution of relevant features over the field~\cite{Galceran:2013kg,10.3390/drones3010004}.

UAV swarms have been proposed to address the adaptive monitoring and mapping of extensive fields. A swarm-robotics approach can indeed improve efficiency thanks to parallel monitoring of the field by different UAVs, reducing the operation to a fraction of the time required by a single-UAV approach~\cite{10.1007/s10846-016-0461-x,Albani:2017cb,Albani:2019eq}. Additionally, accuracy can be improved thanks to collaboration among UAVs in their feature-detection task \cite{magistri-etal:2019imav}. Finally, self-organised deployment strategies can be envisaged to leverage the ability of UAVs to estimate which region within the field is of greater interest (e.g., estimating weed density from high-altitude flights \cite{Lottes:up}) and perform accurate monitoring and mapping only where relevant (e.g., by collecting high-resolution images while flying at low altitude \cite{Albani:2018vj}). With such an approach, UAVs can autonomously decide to enter or leave a given region based on estimates of the monitoring activity's completion level and energy constraints. Consequently, the monitoring and mapping strategy of a single region needs to be flexible and scalable to adapt to changes in the actual number of UAVs that are concurrently operating. Approaches that a-priori divide the field to be monitored among the available UAVs are viable but do not address adaptive deployment requirements into areas of interest.

Improving over previous work~\cite{Albani:2017cb,Albani:2019eq}, we propose a fully decentralised strategy to monitor a region or field based on reinforced random walks (RRW \cite{10.1098/rstb.2010.0078}), which maximises the monitoring effort only on areas that likely provide relevant information. To this end, we exploit the Information Gain (\IG), an information-theoretic measure of the expected reduction in uncertainty from additional observations of a specific area. The usage of information theory for exploration and mapping has been demonstrated across application domains \cite{10.1177/0278364915587723,Ma2018,palazzolo2018effective,ognibene2019proactive,ognibene2013towards}, and precision agriculture in particular \cite{10.1109/icra40945.2020.9196826,Rossello-arxiv}. Here, we exploit \IG to support exploration and coordination among robots. To this end, we build a model of the weed-density uncertainty in a given area that accounts for detection errors of a convolutional neural network (CNN) trained on a real dataset. To enable real-time onboard execution, we reduce the neural classifier's complexity and compensate for increased error by allowing multiple observations of the same area of the field \cite{magistri-etal:2019imav}. On such basis, we compute the expected reduction in uncertainty as to the \IG from repeated independent observations made by UAVs within the swarm on the same area. Then, we exploit \IG to prioritize areas of interest to be observed next and to determine if these areas are likely to be targeted by other UAVs within the swarm. These two aspects are combined to guide the random selection of the next location to visit. Our results with multi-UAV simulations show that the swarm manages to quickly monitor those areas of the field that require more attention, minimising the observation error faster than approaches based on potential fields \cite{Albani:2017cb,Albani:2019eq}, and better than a baseline approach based on a predefined flight plan that uniformly covers the entire field. 
Thus, the main contribution of this research is a parameter-free \IG-based mapping strategy that achieves an improved reduction of observation error during the early stages of inspection, leading to the generation of reliable maps when the time/energy budget may be limited.


\section{Problem description}
\label{sec:problem}

We consider a field-monitoring and weed-mapping problem in which areas of high weed infestation need to be identified, creating a prescription map that can be exploited for weed control (e.g., variable-rate application of herbicides). Specifically, we focus on identifying volunteer potatoes that infest sugar-beet fields---a common benchmark for precision agriculture  \cite{Nieuwenhuizen:2010dg,Lottes:up}---and use automatic object classification to inform the monitoring and mapping strategy. Our goal is to deploy a swarm of UAVs that can rapidly minimize the error in detecting weeds within the field. To this end, we exploit a simulated scenario to evaluate the proposed strategy's effectiveness and scalability.

\subsection{World model and UAV swarm simulation}
We consider the field as divided into small areas forming a grid of cells, and each cell is fully contained within the camera field-of-view of a UAV hovering over its centre. Without loss of generality, we consider here a square field divided in a grid of $C\times C$ square cells, each with side $l_c$. Each UAV travels at a cruise speed of $v=\unitfrac[0.1l_c]{m}{s}$ at an altitude $h$ sufficient to observe the whole cell given the camera footprint (e.g., $h\geq l_c/2$ if the camera aperture is $\frac{\pi}{2}$). Whenever moving over a cell, a UAV takes an RGB image used for crop/weed classification, leading to an estimation of the number of weeds present in the cell (see Section~\ref{sec:model}). After each observation, the UAV updates its local world representation, that is, a $C \times C$ map of the field (see Section~\ref{sec:stats}). Additionally, UAVs can communicate with each other by broadcasting short messages exploiting a radio link, with range $Rl_c$. When the communication range is sufficiently large, any UAV can receive the information shared by any other UAV. 
Otherwise, UAVs apply a simple re-broadcasting protocol to maximise the reach of information shared within the swarm. Upon reception of a message, a UAV re-broadcasts the message once and then puts it in a blacklist hence avoiding overloading the communication channel. UAVs exploit communication to share information about the observations made on visited cells, and to also share their absolute position---available from some GNSS positioning system---hence allowing collision avoidance (here implemented using ORCA~\cite{Bareiss:2015hg}) as well as collaboration for monitoring and mapping.


While crops are uniformly distributed over the field, weeds mostly appear in patches. We consider here $C_p$ patches, each extending on a square of $n_{p} \times n_{p}$ cells, more densely distributed in the center than in the periphery following a Gaussian distribution. Some isolated weeds are also present within $C_i$ additional cells. Overall, the number of cells with some weed is $C_p n_p^2 + C_i$. Each cell can contain at most $N_W$ weeds, i.e., the maximum value observed over the field.

\subsection{Model of weed-classification uncertainty}
\label{sec:model}
\begin{figure}[t]
\centering
\begin{subfigure}{0.4\textwidth}
	\begin{subfigure}{\textwidth}
		\centering
		\includegraphics[width=\textwidth]{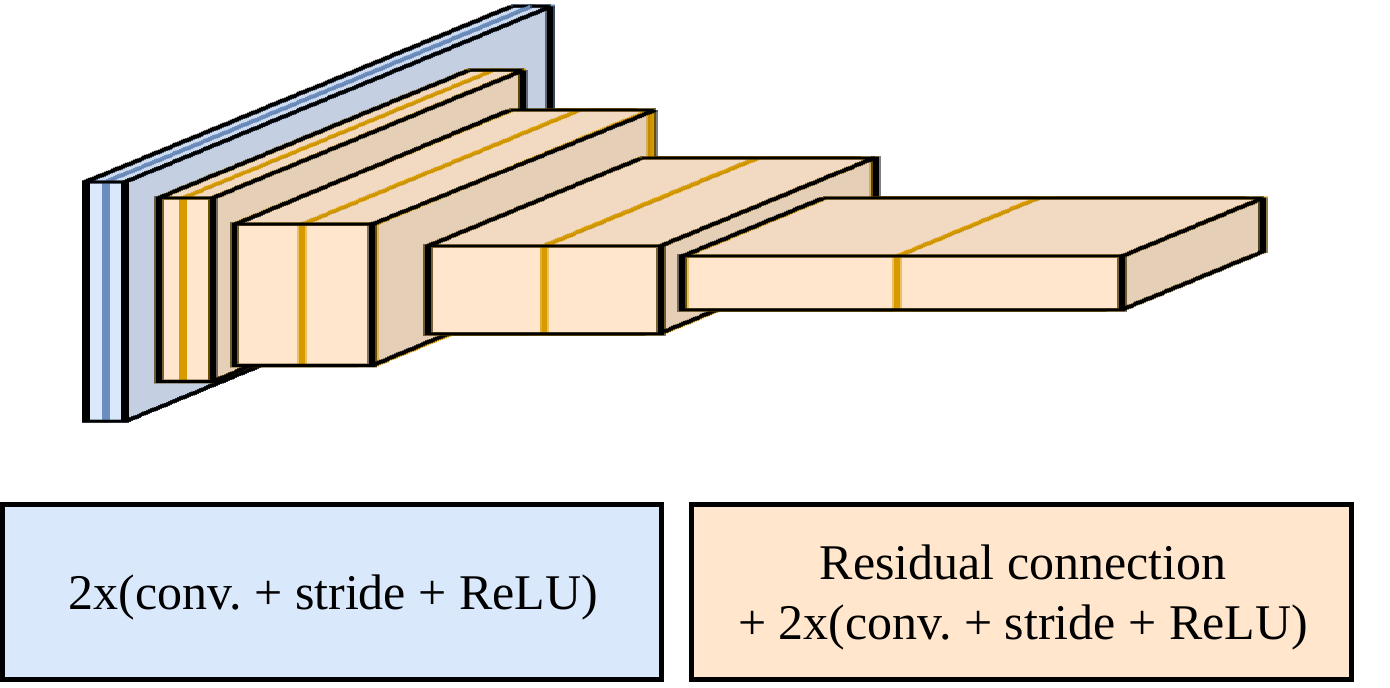}
		\caption{}
		\label{fig:cnn}
	\end{subfigure}%
	\\
	\par\medskip
	\begin{subfigure}{\textwidth}
		\centering
		\includegraphics[width=\textwidth]{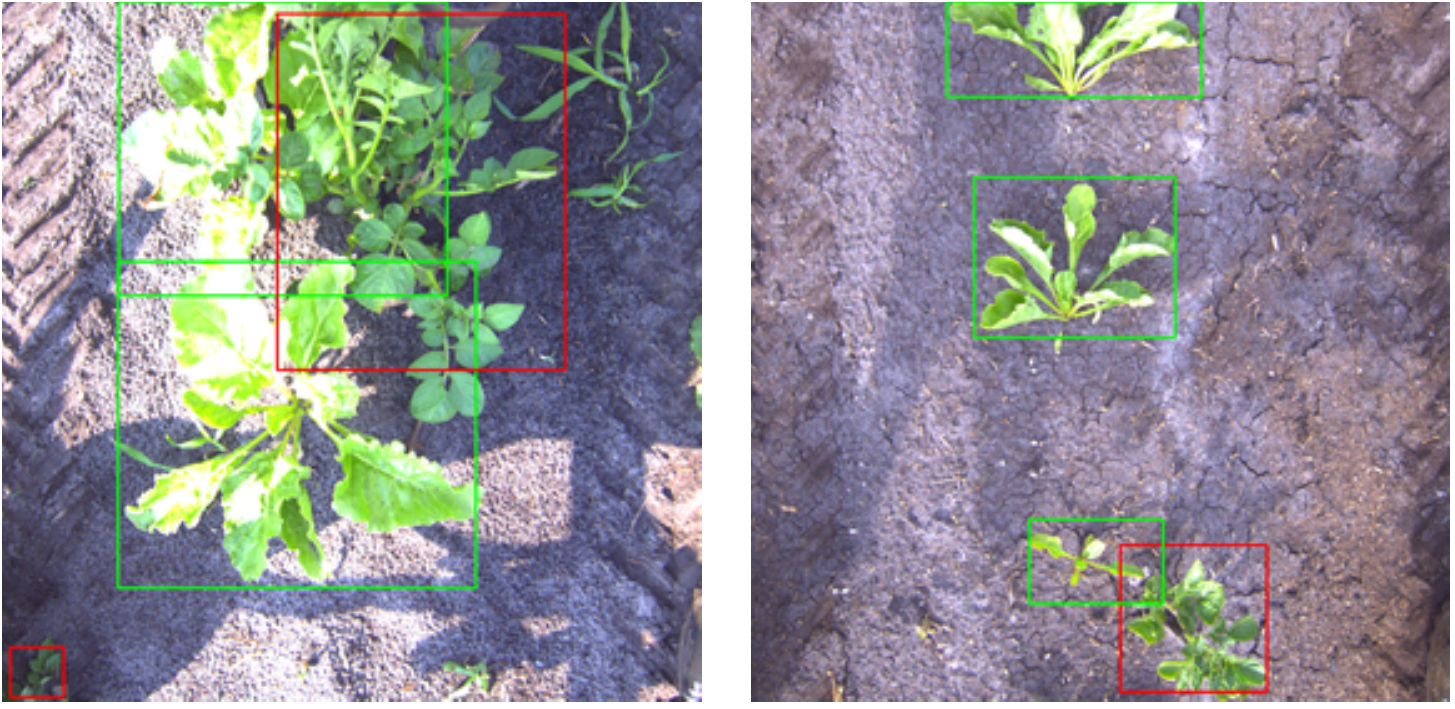}
		\caption{}
		\label{fig:bb}
	\end{subfigure}%
\end{subfigure}
\quad
	\begin{subfigure}{0.5\textwidth}
		\centering
		\includegraphics[width=\textwidth]{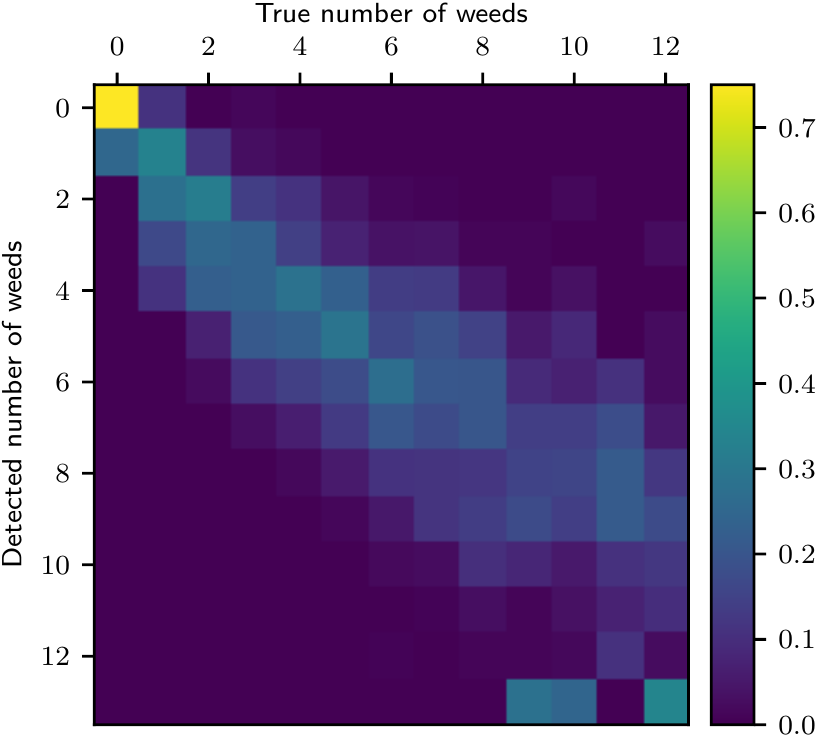}
		\caption{}
		\label{fig:weed_norm}
	\end{subfigure}%
  \caption{(a) The Faster R-CNN framework with a shallow backbone was used while the head of the CNN remained the same as in the original work. (b) a few samples of CNN classification of volunteer potatoes (red) and sugar beets (green). (c) Sensor model derived from the CNN detection performance.}
\label{fig:detections}
\end{figure}

To build a model of the weed-classification uncertainty, we consider a CNN to detect individual plants and label them as crops or weeds. In other words, given an input image, the CNN returns a list of bounding boxes that enclose the detected plants, together with the identified class of the plant. 
We use the state-of-the-art framework \emph{Faster R-CNN}~\cite{faster} but limit its computational requirements by employing a shallow backbone (instead of the deep backbones normally used, such as ResNet-101) to match the constraints imposed by the limited power available on UAVs (see Figure~\ref{fig:cnn}). The input image is passed through a convolutional block followed by 4 residual blocks~\cite{he2016cvpr}. The convolutional block is composed of a $7\times7$ convolutional layer followed by a $3\times3$ convolutional layer (represented in light blue in Figure~\ref{fig:cnn}), each layer having 64 filters with stride 2 and ReLU activation functions. Each residual block is composed of a residual connection and two $3\times3$ convolutional layers with ReLU activation functions (represented in orange in Figure~\ref{fig:cnn}). The first residual block has 64 filters. Afterward, each block doubles the previous number of filters. The head of the network is the same as in the original work, predicting the location, size, and class of object detections. We exploit a dataset collected at Wageningen University, composed of 500 images---400 used as the training set and 100 as the test set---labelled to represent sugar beets in the crop class and volunteer potatoes in the weed class (see Figure~\ref{fig:bb}). We train our network for 10,000 epochs, using the ADAM optimizer~\cite{kingma2014adam} with a learning rate of 0.01. We evaluate our CNN using the average precision (AP) metric as defined in the MS-COCO challenge~\cite{lin2014eccv} obtaining an AP score of 89.6 for the crop and 56.1 for the weed. While the AP for the crop class is in line with the literature~\cite{fawakherji2019irc,milioto2017isprs}, the lower accuracy for the weed class is mainly due to miss-classifications of plants as background (Figure~\ref{fig:bb}). 

We model the uncertainty in the weed classification to obtain a realistic observation---in terms of detected weeds---every time a UAV processes the information of a cell, as well as to update the internal knowledge about the world, as discussed in the following. 
Starting from the trained CNN classification output, our model associates a probability distribution to all possible observations given the actual number of weeds present in a cell. Considering that each cell can contain a discrete and small number of weeds (from the available data, we estimated $N_W=12$ weeds for $l_c=\unit[4]{m}$), we build a table that associates the actual number of weeds $w$ with each possible observed number $o$, and gives for each combination a probability of occurrence $P(o|w)$. This table is estimated from the available data exploiting the trained CNN. Specifically, we compare the number of detected weeds against the true number of weeds. The element $T(o,w)=P(o|w)$ of the table is the relative frequency of detecting $o$ weeds when the true number is $w$. Having several false positive cases, we also consider into the table the case of detecting more weeds than the given maximum $N_W$, that is, $o\geq N_W+1\triangleq N_W^+$ represents the case in which the number of weeds detected is larger than $N_W$. The trained CNN model as obtained from the available data is presented in Figure \ref{fig:weed_norm}. Both false positives and false negatives are possible, with errors becoming more frequent when the number of weeds is larger, as one would expect.

\subsection{Uncertainty reduction from multiple observations}
\label{sec:stats}
For each cell $c$, each UAV maintains a knowledge vector $p_c = [p_c(0),...,p_c(N_W)]$ that represents the expected probability distribution of all possible number of weeds considered. Taking a conservative approach, when no observation for a cell has been performed and no prior knowledge is available, the probability distribution is assumed to be uniform, with every element $p_c(w)=\frac{1}{N_W+1}$.  
Whenever a UAV observes cell $c$ at time $t$, it detects a number of weeds $o_c^t$ that depends on the actual value of weeds $w_c$ present in $c$. Following the observation $o_c^t$, the probability $P_c(w|o_c^t)$ represents the updated knowledge vector for each possible value $w$. This can be easily computed exploiting Bayes' theorem:
\begin{equation}
    \label{eq:bayesKV}
    \forall w, p_c(w) \leftarrow P_c(w|o_c^t ) = \frac{p_c(w) P(o_c^t|w)}{P(o_c^t)} = \frac{p_c(w) T(o_c^t,w)}{\sum_{j=0}^{N_W} p_c(j) T(o_c^t,j)}
\end{equation}
The residual uncertainty about the weeds present in the cell corresponds to the information entropy of the knowledge vector:
\begin{equation}
    \label{eq:residualUncertainty}
    H_c(W) = \sum_{w=0}^{N_W}-{p_c(w)\log(p_c(w))}
\end{equation}
The residual uncertainty is updated at every additional observation. We consider that sufficient observations have been performed for a given cell $c$ when the entropy decreases past a low threshold that we heuristically set at $\hat{H} = -\frac{N_W-1}{N_W}\log\frac{N_W-1}{N_W} - \frac{1}{N_W}\log\frac{1}{N_W}$, which corresponds to the case in which there are only two remaining alternatives, one of which is much more likely than the other. When the threshold is reached, cells are marked as mapped and are not considered anymore for further observations. Note that other UAVs can also share observations, allowing to update the residual uncertainty about a cell also when others have visited it. In this way, UAVs try to maintain aligned their local representation of the field. 

The residual uncertainty of a cell is also the first component for the calculation of the \IG of a cell, which represents the expected reduction in entropy from any possible additional observation $o$. 
A UAV can compute the \IG of a cell $c$ 
as follows:
\begin{equation}
\label{eq:IG}
\IG_c(W) = H_c(W) - H_c(W|O)
\end{equation}
where $H_c(W|O)$ is the conditional entropy of the same cell given that additional observations will be performed next:
\begin{align}
\label{eq:conditionalEntropy}
H_c(W|O) =& 
-\sum_{o=0}^{N_W^+}P_c(o)\sum_{w=0}^{N_W}\left[P_c(w|o)\log\left(P_c(w|o)\right)\right]=\\
=&-\sum_{o=0}^{N_W^+}\left[\sum_{j=0}^{N_W}T(o,j)p_c(j)\right] \sum_{w=0}^{N_W}\left[P_c(w|o)\log\left(P_c(w|o)\right)\right]
\end{align}
Hence, based on the available knowledge, each UAV can compute the \IG for each cell of the field and quantify the information gathered from a new observation, that is, the utility of visiting it.

\section{Reinforced random walks for monitoring and mapping}
\label{sec:motion_planning}
A RRW is an exploration strategy that exploits available information to bias the random selection of the targets \cite{10.1098/rstb.2010.0078}. The available information can derive from world knowledge (e.g., avoid visiting cells that are marked as mapped) or from other UAVs (e.g., others' location to avoid interference). A wise combination of both aspects can produce efficient monitoring and mapping strategies that focus on relevant areas, minimising the detection error.

\subsection{Neighbourhood selection strategy}

Whenever a UAV $i$ has to determine the next cell to visit, it considers only those in its neighbourhood $\mathcal{N}_i$ as targets, which proved to be the best strategy to avoid long relocations that may be costly \cite{Albani:2017cb,Albani:2019eq}. Additionally, a UAV tries to select cells that are \emph{valid}, i.e., not yet marked as mapped and not currently targeted by other UAVs to avoid interference.
We define the neighbourhood $\mathcal{N}_i^d$ of UAV $i$ as the cells at Chebyshev distance $d$. A UAV first considers the neighbourhood $\mathcal{N}_i^1$, and only if there are no valid cells available, it considers the neighbourhood $\mathcal{N}_i^2$. If this does not contain valid cells, the UAV selects a cell in $\mathcal{N}_i^2$ to move away from the current position, limiting the choice among those cells that are already marked as mapped. With this strategy, a set of neighbouring cells $\mathcal{V}_i$ is selected for a decision to be taken.
This strategy, being purely local, does not guarantee that all cells are eventually visited, even if, in the long run, this can always happen in practice. Nevertheless, in this work, we focus on reducing the mapping error with a limited time/energy budget. Hence complete coverage is not a requirement.

\subsubsection{\IG-based RRW}
The cell selection to target is based on a utility measure computed starting from the \IG. Given a cell $c_k$, each UAV $i$ determines the $\IG^i_{c_k}$ based on the currently available knowledge---different among UAVs in case of constrained communication---and, independently from any other UAV, assigns a probability of selecting the cell proportional to the relative \IG:
\begin{equation}
\label{eq:probOnlyIG}
P^\IG_{i,{c_k}} = \frac{\IG^i_{c_k}(W)}{\sum_{z\in\mathcal{V}_i}{\IG^i_{c_{z}}(W)}}
\end{equation}
To take into account the presence of other UAVs that are concurrently monitoring the field, the utility of selecting a cell is computed from $P^\IG_{i,{c_k}}$ considering the likelihood that the cell is \emph{not} chosen by other UAVs:
\begin{equation}
\label{eq:probIGSocial}
u_{i,c_k} = P^\IG_{i,{c_k}}\prod_{j \neq i}{\left [ 1-P^\IG_{j,{c_k}}\right]}
\end{equation}
Here, $P^\IG_{j,{c_k}}$ is computed considering the neighbourhood $\mathcal{V}_j$ of UAV $j$, although such computation is performed exploiting $i$'s private knowledge. To limit the computational complexity of this calculation, the product is extended only to those UAVs $j$ that could potentially target the cell $c_k$ at the same time, that is, the UAVs within the neighbourhood $\mathcal{N}^d_{c_k}, d\le2$ from cell $c_k$. By reducing the likelihood of choosing cells in reach from other UAVs, the proposed RRW strategy implements an implicit coordination mechanism that allows UAVs to share the monitoring burden and efficiently map the relevant areas. Should UAVs decide to target the same cell, the ORCA method would prevent collisions, and a timeout mechanism allows to resolve possible---albeit unlikely---deadlocks.
Given eq.~\eqref{eq:probIGSocial}, we propose different heuristics to choose the next cell to visit:
\begin{description}[align=right,labelwidth=4ex,topsep=0pt]
\item[G:] a cell $c_k$ is chosen greedily selecting the one with highest utility $u_{i,c_k}$. In case of cells with identical utility, one is chosen at random. 
\item[R:] a cell $c_k$ is randomly chosen proportionally to the utility $u_{i,c_k}$.
\item[S$_\gamma$:] a cell $c_k$ is chosen according to a softmax function of $u_{i,c_k}$ with base $e^\gamma$.
\end{description}

\subsection{Baseline strategy based on optimal pre-planned trajectories}
 As a baseline to confront the proposed strategy, we assume an optimal reference point $\mathbf{B}$ based on the uniform coverage of the field, i.e., $N$ UAVs visit each cell of the field a fixed number of times. This optimal benchmark could be approximated by pre-planning all UAVs' trajectories, although, in reality, several factors reduce the ideal performance proposed here. 
 Considering the speed of the UAV and the distance between cells, the time for a single UAV to fully cover the field is given by $\mathcal{T}_1= C^2l_c/v = 10C^2$. We assume that $N$ UAVs can optimally partition the field, hence $\mathcal{T}_N = \mathcal{T}_1/N$. To allow for independent observations of the field by different UAVs, we consider repeated passages that entail longer times, that is, $M\mathcal{T}_N$ for $M$ independent passages. After each observation, the residual uncertainty of a cell $c$ is computed following eq.~\eqref{eq:residualUncertainty}.

\subsection{Baseline strategy based on potential fields}
In addition, we consider a second baseline $\mathbf{B_{PF}}$ that exploits a RRW based on potential fields (PF). Here, the target cell is selected by UAV $i$ according to a directional bias given by an attraction vector $\vec{a}_i$ toward areas where weed was detected, and a repulsion vector $\vec{r}_i$ from other agents to avoid overcrowding. This strategy is adapted from previous versions~\cite{Albani:2017cb,Albani:2019eq} to reduce its computational demands, limiting the information exploited for the computation of potential fields. Attraction and repulsion vectors are computed as follows: %
\begin{equation}
  \label{eq:1}
  \vec{r}_i = \sum_{j\neq i} S(\vec{x}_i-\vec{x}_j, \sigma_r),\;
  \vec{a}_i = \sum_{c} \frac{\hat{w}_c}{N_W} S(\vec{x}_c-\vec{x}_i, \sigma_a),\;
    S(\vec{v},\sigma)=2 e^{\mathrm{i}\angle\vec{v}} e^{-\frac{|\vec{v}|}{2\sigma^2}},
\end{equation}
where  $\vec{x}$ represents the position of an agent/cell, and
$S(\vec{v},\sigma)$ returns a vector in the direction of $\vec{v}$
with a Gaussian length with spread $\sigma$. With respect to
\cite{Albani:2019eq}, we reduce the number of agents considered to those belonging to the neighbourhood $\mathcal{N}^4_i$, which are ultimately considered also for the \IG-based RRW. Additionally, we consider attraction to the cells $c\in\mathcal{N}^d_i,d\le2$ but discounting the force by the number $\hat{w}_c$ of weeds detected in the last observation. Cells are not considered for attraction when they are marked as mapped.
The selection of the next cell to visit by UAV $i$ is performed randomly using the vector $\vec{v}_i=\vec{r}_i+\vec{a}_i$ as a bias. Specifically, the cell $c_k\in\mathcal{V}_i$ is selected randomly proportionally to the utility $u_{i,c_k}$, which
is computed according to the angular difference $\theta_{i,c_k} = \angle (\vec{x}_{c_k} - \vec{v}_i)$:
\begin{equation}
  \label{eq:cauchy}
  u_{i,c_k} = C\left(\theta_{i,c_k},0.9\left(1-e^{\beta|\vec{v}_i|}\right)\right),\quad C(\theta,p) =\frac{1}{2\pi}
  \frac{1-p^2}{1+p^2-2p\cos \theta}
\end{equation}
where $C(\cdot)$ is the wrapped Cauchy density function with persistence $p$. In this way, the length of the vector---smoothed through an exponential ceiling---determines the relevance according to the bias: the smaller the module, the lower the directional bias in the cell choice. This strategy has several parameters---namely the Gaussian parameters $\sigma_a$ and $\sigma_r$, and the exponential constant $\beta$---which depend on the UAV swarm size $N$ and need to be carefully tuned for appropriate performance~\cite{Albani:2017cb,Albani:2019eq}.

\section{Experimental Results}
\label{sec:results}

We performed several simulations with a square field of $C \times C=2500$ cells, having $C_q=4$ square weed patches of $n_p \times n_p=49$ cells randomly positioned within the field, plus additional $C_i=40$ isolated cells randomly scattered. In total, less than 10\% of the field presents cells containing weeds. As already mentioned, we consider $N_W=12$. At initialisation, UAVs start at random positions within the field, with no prior knowledge of the weed distribution.\footnote[1]{We ignore here the initial relocation from a deployment station and also disregard the need to return to a predefined location.} We focus on the ability to monitor and map the field minimising the mapping error. We perform 50 runs for each experimental condition obtained varying swarm size $N\in\{10,20,30,40,50\}$, the communication range $R\in\{10,\infty\}$, the heuristics exploited for the information gain ($\mathbf{G}$, $\mathbf{P}$ and $\mathbf{S_\gamma}$ with $\gamma\in\{1,5\}$) and the parameters $\sigma_r,\sigma_a\in\{0,2,4,8,16,32\}, l_c$ for the PF-based RRW with $\beta=1$. With ideal communication ($R=\infty$), each broadcasted message reaches every other UAV in the swarm. 

\begin{figure}[t]
  \centering
  \includegraphics[width=\textwidth]{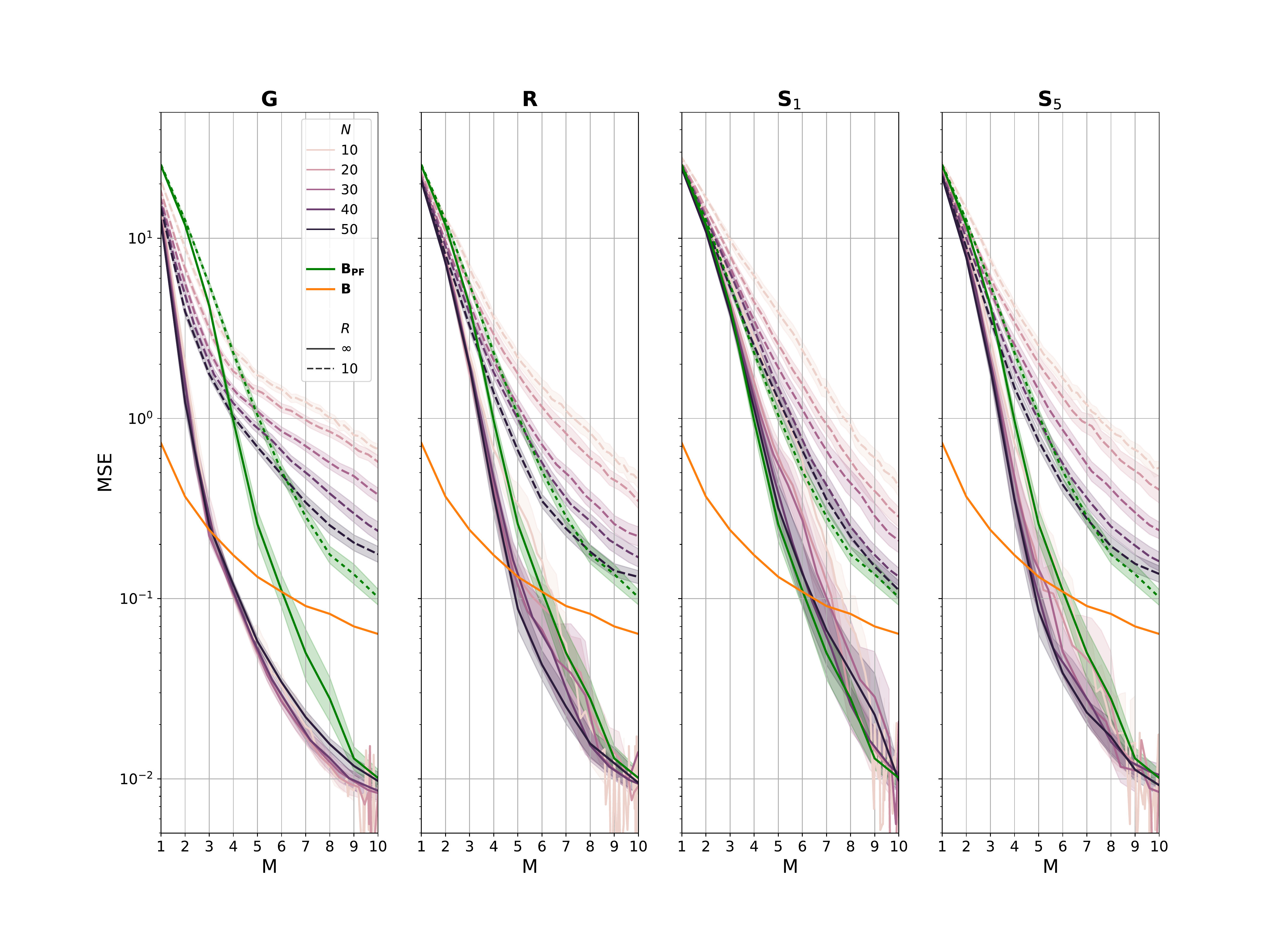}
  \caption{MSE of the maps generated with different strategies (average and standard deviation), with respect to the optimal pre-planned monitoring baseline $\mathbf{B}$ and the PF-based RRW strategy $\mathbf{B_{PF}}$. Note that all plots are rescaled in time with respect to the time $\mathcal{T}_N$ required by $N$ agents to fully cover the field once. As a consequence, the average MSEs from the baseline $\mathbf{B}$ for all values of $N$ coincide. The baseline $\mathbf{B_{PF}}$ is shown for both $R=\infty$ and $R=10$, and corresponds to the case with $\beta=1$, $\sigma_a=2$, $\sigma_r=8$ and $N=50$, which is the one with lowest MSE among all the tested parameters.}
  \label{fig:mse_all}
\end{figure}

To evaluate the system performance, we consider an aggregated world map resulting from the individual UAV maps by considering for each cell $c$ and UAV $i$ the knowledge vector $p_c = p_{c,i}$ with lowest uncertainty $H_i(c)$. Then, the value $\tilde{w}_c = \arg\max_w p_{c}$ is considered as the mapped number of weeds in cell $c$, and the mean squared error (MSE) is computed with respect to the real value $w_c$ for the whole field. Figure~\ref{fig:mse_all} shows how this error decreases as further observations are gathered from the field, taking as temporal reference the time $M\mathcal{T}_N$ necessary to gather $M$ independent observations for each cell with $N$ UAVs following the baseline $\mathbf{B}$. When the communication is perfect, all UAVs share the same map, and any new observation contributes to reducing the uncertainty. The mapping error is initially rather high but decreases as UAVs discover and focus on interest areas. 
Notably, all the \IG-based strategies scale well to the swarm size $N$ and present the best performance when the greedy approach $\mathbf{G}$ is employed. In this case, the MSE gets better than the one of the baseline $\mathbf{B}$ for $M\approx3$---when the error is about 0.15---and continues to decrease another order of magnitude with further observations. The baseline $\mathbf{B_{PF}}$ is instead less efficient in reducing the mapping error and outperforms $\mathbf{B}$ only when $M>6$, hence requiring much longer than the best IG-based strategy $\mathbf{G}$. The other heuristics perform slightly worse, meaning that additional randomness in selecting the next cell to visit does not increase performance. Specifically, the $\mathbf{R}$ and the $\mathbf{S}_5$ strategies have very similar profiles, and the worst performance is obtained with the $\mathbf{S}_1$ strategy.

\begin{figure}[t]
  \centering
  \includegraphics[width=\textwidth]{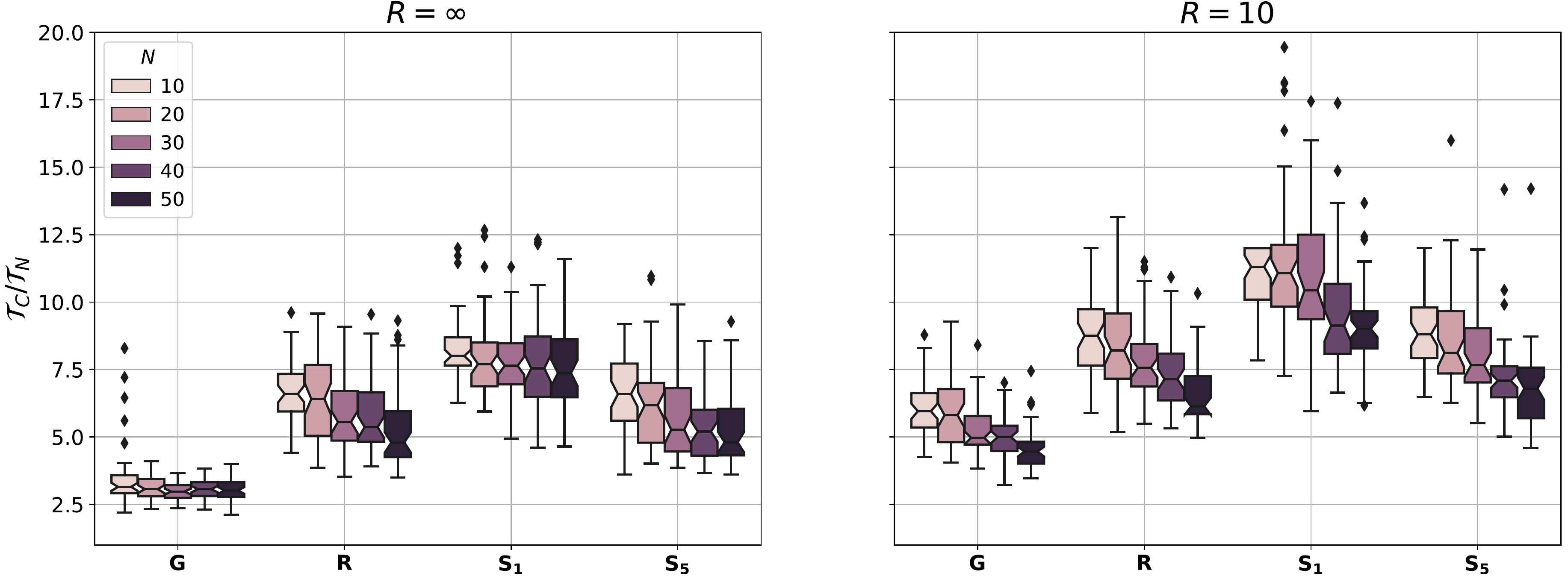}
  \caption{Coverage time $\mathcal{T}_C$ relative to $\mathcal{T}_N$ for the different IG-based strategies, for $R=\infty$ (left) and $R=10$ (right).}
  \label{fig:time_all}
\end{figure}

The difference in performance among the proposed heuristics is also visible for the total completion times, shown in Figure~\ref{fig:time_all}. Here, we measure the time $\mathcal{T}_C$ necessary to fully cover the entire field by passing over every cell at least once, scaled concerning the time $\mathcal{T}_N$ of the baseline $\mathbf{B}$. The greedy strategy $\mathbf{G}$ performs better than any other strategy, with a coverage time that scales perfectly with group size and values that roughly correspond to the time required to have an MSE smaller than $\mathbf{B}$.  The other strategies generally have longer coverage times, with slight improvements for larger group size $N$ although never being on par with $\mathbf{G}$.

\begin{figure}[t]
  \centering
  \includegraphics[width=\textwidth]{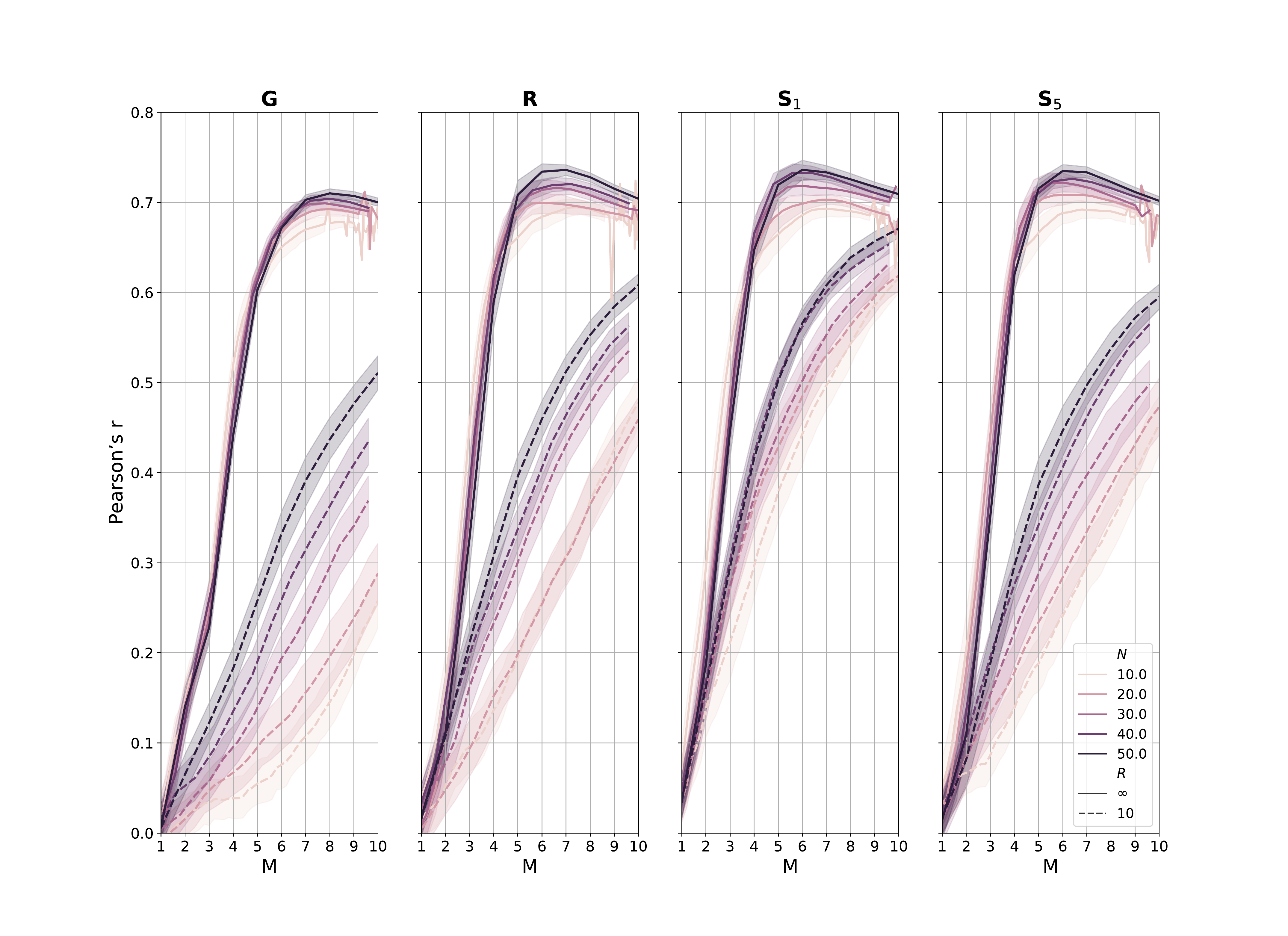}
  \caption{Correlation between the number of weeds present and the number of observations performed, plotted over time for the different strategies.}
  \label{fig:correlation_all}
\end{figure}

The adaptive approach reduces the uncertainty below the optimal pre-planned strategy by mainly focusing on the areas where weed is present, avoiding monitoring those devoid of weeds. Indeed, Figure~\ref{fig:correlation_all} shows the Pearson's correlation coefficient between the number of weeds present in a cell and the number of independent observations gathered for it. It is possible to note that, as time goes by, the relationship between these two variables builds stronger, meaning that more time is spent over areas that require more observations to reduce the uncertainty substantially. In contrast, areas without weed are quickly abandoned.

The results obtained for a limited communication range tell a different story, however. First and foremost, the swarm does not reduce the error substantially below the baseline $\mathbf{B}$ (see dotted lines in Figure~\ref{fig:mse_all}). The monitoring and mapping performance is substantially worse due to an inefficient reduction of uncertainty as the observations from other UAVs are only shared locally. There is no mechanism to ensure that UAVs maintain the local maps aligned. The error gets lower as the group size increases due to the larger diffusion of messages thanks to the re-broadcasting communication protocol. Also, the coverage time is positively affected by group size, as shown in Figure~\ref{fig:time_all}. Nevertheless, the density of agents is not sufficiently high to ensure that the communication topology is always connected, leading to a partial loss of communications. Consequently, different UAVs may map areas already sufficiently observed by others without any improvement for the collective. This also leads to a reduced correlation between the number of weeds in a cell and the number of observations, especially for low values of $N$ as shown in Figure~\ref{fig:correlation_all}.

\section{Conclusions}
\label{sec:conclusions}

This paper has demonstrated that a performance improvement is expected when adaptive approaches are employed, leading to a substantial reduction of uncertainty below what can be achieved through a blind acquisition of information using pre-planned trajectories. The mapping error is substantially reduced by investing a little more exploration time despite the low accuracy in the classification output of single images. Additionally, the best heuristic based on \IG is rigorous. It bears no free parameters, being therefore ideal for deployment in a swarm robotics system notwithstanding the group size or the field dimensions, as no tuning is needed.  

One observed limitation concerns the loss in performance for constrained communication, which jeopardises the benefits from an adaptive approach with a UAV swarm.
This aspect is worth studying in detail, although it does not invalidate the ideal communication case results. In several scenarios---and precision agriculture in particular---one can reasonably assume that a good-enough communication channel can be set up in place, thanks to long-range radio communication or the upcoming 5G technology.
The limited communication scenario may be addressed by either changing the communication protocol in a way to ensure better alignment among UAVs of the local world representations (e.g., by sharing not just the latest observation but the entire word representation), or by explicitly maintaining a high degree of connectivity within the swarm.

Another limitation of the proposed approach stands in the assumption of independence between observations, which likely does not hold in real-world scenarios, especially for slowly-changing environmental conditions as in crop fields. Indeed, multiple observations are likely to provide the same errors if these are related to specific features of the observed area (e.g., a volunteer potato hiding between two sugar beets), especially if observations are made from the same perspective a short time-frame. To mitigate this issue and increase the chances that repeated observations lead to substantial uncertainty reduction, a possible approach is to vary the observation position relative to the point of interest. Different perspectives can lead to more informative observations. Another complementary approach consists of using already available knowledge as a prior for the classification, exploiting specially-trained CNNs \cite{magistri-etal:2019imav}. Combining these two possibilities can lead to a substantial reduction in uncertainty that can be modelled and exploited with the approach proposed in this paper. Future work will investigate this possibility while tackling real-world monitoring and mapping tasks.

\section*{Acknowledgements}
This work has partially been funded by the Deutsche Forschungsgemeinschaft (DFG, German Research Foundation) under Germany's Excellence Strategy, EXC-2070 -- 390732324 (PhenoRob). Vito Trianni acknowledges partial support from the project TAILOR (H2020-ICT-48 GA: 952215).

\bibliographystyle{abbrv}
\bibliography{templates/Swarm_01}

\end{document}